\documentclass[conference]{IEEEtran}
\IEEEoverridecommandlockouts

\usepackage{cite}
\usepackage{amsmath,amssymb,amsfonts}
\usepackage{algorithmic}
\usepackage{graphicx}
\usepackage{textcomp}
\usepackage{xcolor}
\usepackage{booktabs}
\usepackage{multirow}

\def\BibTeX{{\rm B\kern-.05em{\sc i\kern-.025em b}\kern-.08em
    T\kern-.1667em\lower.7ex\hbox{E}\kern-.125emX}}
\begin{document}

\title{Closed-Loop Hybrid Digital Twin Platform for Connected and Automated Vehicle Validation}

\author{  
    \IEEEauthorblockN{Kanglong Quan, Zhebing Xia, Linfeng Jiang, Hao Yu, Ziheng Qiao, Dapeng Dong, and Dongyao Jia\IEEEauthorrefmark{1}}
    
    \IEEEauthorblockA{School of Advanced Technology\\
    Xi'an Jiaotong-Liverpool University, Jiangsu, China\\
    Email: \{Kanglong.Quan24 , Zhebing.Xia22, Linfeng.Jiang22,\\ Hao.Yu24, Ziheng.Qiao18\}@student.xjtlu.edu.cn, \{Dapeng.Dong, Dongyao.Jia\}@xjtlu.edu.cn}

\thanks{This work is supported in part by the National Natural Science Foundation of China under Grant 62372384, and in part by the Suzhou Science and Technology Development Planning Programme under Grant ZXL2024342. (\textit{Corresponding author: Dongyao Jia)}}
    
}

\maketitle

\begin{abstract}
Comprehensive and efficient validation of connected and automated vehicles (CAVs) is critical prior to real-world deployment. While simulation-based testing offers scalability, existing approaches often lack seamless integration with real vehicles and field data, limiting their fidelity in capturing dynamic, real-world interactions. To bridge this gap, this paper proposes a novel real-time hybrid digital twin platform. Its core innovation lies in the tight coupling of a high-fidelity CARLA-SUMO co-simulation with a physical test site and vehicle via a low-latency Vehicle-to-Everything (V2X) communication link. A custom-developed middleware serves as the critical bridge, synchronizing a real CAV's kinematic state as a shadow vehicle in the simulation and translating virtual control commands into chassis-actuating Controller Area Network (CAN) messages for closed-loop control. Detailed implementation includes using photogrammetry for full-scale asset reconstruction and a cloud-edge collaborative architecture for scalable, multi-user operation. Experimental results demonstrate stable synchronization and effective closed-loop control with low latency, confirming the platform's practicality for multi-scenario CAV verification.

\end{abstract}

\begin{IEEEkeywords}
Hybrid digital twin, V2X communication, CAVs
\end{IEEEkeywords}

\section{Introduction}
Connected and Autonomous Vehicles (CAVs) and intelligent transportation systems are rapidly evolving toward large-scale deployment, which places increasingly stringent requirements on testing and verification. However, validating CAV functions in real-world traffic remains challenging. On-road experiments are expensive \cite{zhao2017accelerated}, safety-critical, and difficult to reproduce due to uncontrollable environmental factors and complex interactions among vehicles and infrastructures. Although simulation-based testing offers controllability, repeatability, and scalability \cite{JIA2021102984,sasaki2023proposal}, such purely virtual approaches often fail to capture key characteristics of the real-world, such as vehicle dynamics, sensing noise, and unreliable communication links \cite{cheng2024survey,daza2023sim,cui2022cooperative}. Consequently, a digital twin-assisted testing paradigm is becoming attractive: it preserves the controllable and repeatable nature of virtual environments while introducing real vehicles and real Vehicle-to-Everything (V2X) communication to form data-in-the-loop, closed-loop validation \cite{dong2023mixed}\cite{wu2025digital}.

Despite advances in mixed-reality approaches, existing frameworks still suffer from significant limitations. Current digital twin studies focus primarily on traffic monitoring or high-level advisory feedback (e.g., speed guidance) rather than direct vehicle control. They typically rely on simplified vehicle dynamic models in simulation, lacking the fidelity to capture real chassis responses such as non-linear steering actuation and turning rates which are critical for safety-related verification. Moreover, many solutions simulate V2X communication delays purely in software, failing to incorporate real hardware and thus neglecting device-level constraints like packet loss and processing latency under heavy communication loads. These gaps motivate a practical framework that not only bridges the physical and virtual domains but also enables closed-loop chassis control with real V2X hardware-in-the-loop, ensuring credible simulation-to-reality transfer.

This paper proposes a real-time hybrid digital twin platform for CAVs, integrating the CARLA \cite{dosovitskiy2017carla} and SUMO \cite{krajzewicz2002sumo} simulators with a physical vehicle via a low-latency V2X link using On-Board (OBU) and Roadside (RSU) units. To maintain consistency between the physical and digital worlds, the vehicle's kinematic states are transmitted to the simulation, where they control a corresponding \textit{shadow} vehicle. A middleware layer, deployed on a workstation, bridges the two domains via socket-based interfaces. This middleware updates the shadow vehicle's state and, in the reverse direction, converts upstream control commands into Controller Area Network (CAN) frames according to the vehicle's communication matrix, enabling closed-loop chassis control. The main contributions of our work are summarized as follows: 
\begin{itemize}
\item We propose a real-time hybrid digital twin platform that integrates CARLA and SUMO to synchronize a physical vehicle into a mixed-reality traffic environment, enabling high-fidelity, reproducible CAV verification.
\item We develop a low-latency physical–digital bridging middleware, which provides socket-based message exchange, maintains the shadow vehicle via kinematic injection, and generates communication-matrix-compliant CAN commands for closed-loop chassis control.
\item The whole platform is prototyped and deployed in a cloud-edge collaborative testing mode that offloads heavy computation and rendering to a leader edge workstation while allowing lightweight web-based access for multi-user interaction, improving practicality and scalability for multi-scenario testing.
\end{itemize}

\section{Related Works}

\subsection{Simulation-Only Co-Simulation Platforms}
Simulation-only testbeds provide controllability and reproducibility and have been widely used to evaluate CAV functions in scalable traffic scenarios. Representative efforts include integrated traffic-vehicle co-simulation frameworks that couple microscopic traffic simulation (e.g., SUMO) with high-fidelity vehicle simulation (e.g., CARLA) to support cooperative driving scenarios beyond single-vehicle testing \cite{shi2022integrated}. SimCCAD  \cite{JIA2021102984} further designed an integrated platform that combines traffic networks, V2X communication, and vehicle models by integrating SUMO with network simulation (OMNeT++) and robotics/vehicle simulation (Webots) under a client-server deployment model. 

While these simulation-only platforms are strong in traffic realism and virtual vehicle or sensor realism, their overall fidelity is fundamentally bounded by modeling assumptions. In particular, they typically lack real chassis response and actuation delays,  real sensor noise and disturbances, and hardware-level V2X characteristics such as device constraints and link behaviors observed in field deployments. 
\subsection{Digital Twin and Mixed-Reality Testbeds}
To improve fidelity beyond purely virtual testing, recent studies have explored digital twin and mixed-reality paradigms by incorporating real-world data and infrastructure into the simulation loop. Wu et al. \cite{wu2025digital} developed a digital twin framework for V2X-enabled connected vehicle corridors, leveraging real-world C-V2X infrastructure and multi-sourced data to replicate vehicle behaviors, signal timing, communication, and traffic patterns in a virtual environment, with robust synchronization and demonstrated feedback applications. In parallel, Dong et al. \cite{dong2023mixed} proposed the Mixed Cloud Control Testbed (MCCT) under the notion of mixed digital twin, integrating physical, virtual, and mixed experimental platforms, where a cloud unit fuses information and dispatches control instructions to enable synchronous cross-platform interaction and multi-vehicle coordination. 

Although these approaches significantly enhance fidelity through physical-virtual integration, they may still face practical limitations for real-time CAV validation, such as latency introduced by centralized/cloud fusion, limited control capability toward real vehicle actuation, or the lack of an engineering-ready bridge that supports both high-fidelity simulation and hardware-in-the-loop closed-loop chassis control, as summarized in Table~\ref{tab:feature_comparison}. These gaps motivate a hybrid digital twin framework that couples CARLA-SUMO co-simulation with low-latency V2X hardware links and a workstation-deployed middleware enabling real-time shadow vehicle synchronization and closed-loop vehicle control.

\begin{table}[t]
\centering
\caption{Feature Comparison of Mainstream Framework and Ours}
\label{tab:feature_comparison}
\renewcommand{\arraystretch}{1.15}
\resizebox{\columnwidth}{!}{%
    \begin{tabular}{lcccc}
    \hline
    \textbf{Feature} & \textbf{SimCCAD}\cite{JIA2021102984} & \textbf{MCCT}\cite{dong2023mixed} & \textbf{Wu}\cite{wu2025digital} & \textbf{Ours} \\
    \hline
    Scalability \& Reproducibility      & \checkmark & \checkmark & \checkmark & \checkmark \\
    Traffic-Vehicle Co-Simulation          & \checkmark & \checkmark & \checkmark & \checkmark \\
    Real Vehicle Sync                       &            & \checkmark & \checkmark & \checkmark \\
    V2X Hardware-in-the-Loop           &            &            & \checkmark & \checkmark \\
    Closed-Loop Chassis Control             &            & \checkmark &            & \checkmark \\
    \hline
    \end{tabular}%
}
\vspace{-10pt}
\end{table}

\begin{figure*}[t] 
    \centering
    \includegraphics[width=0.9\textwidth]{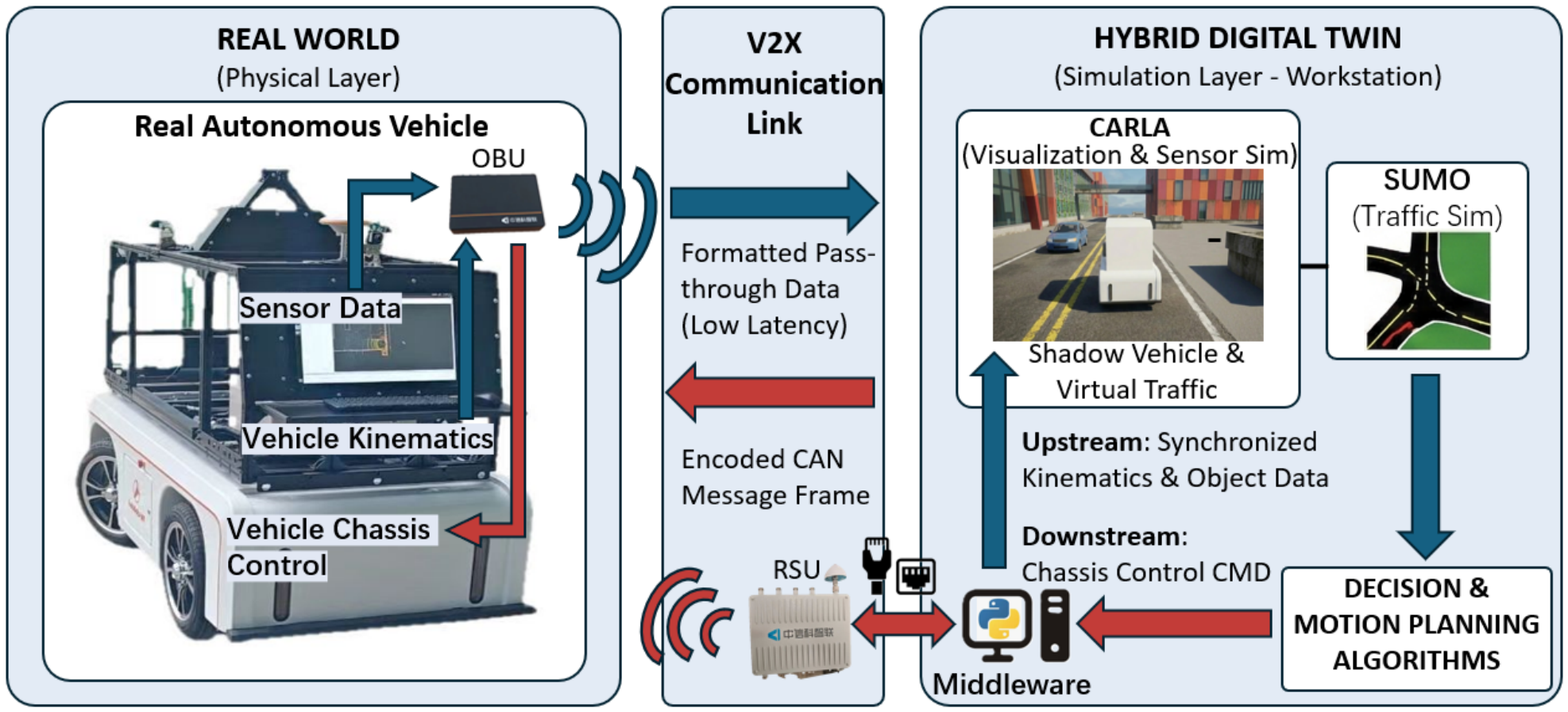}
    \caption{Overview of the hybrid digital twin framework. }
    \label{fig:archi}
    \vspace{-10pt}
\end{figure*}

\section{Real-time Hybrid Digital Twin Framework}

\subsection{ Framework Overview}
As shown in Fig.~\ref{fig:archi}, the whole framework comprises three components: the Physical Layer, the V2X Communication Link, and the Hybrid Digital Twin Layer. The physical layer includes a real test site and a vehicle equipped with an on-board computer and multi-modal sensors (e.g., LiDAR/radar and surround-view cameras), enabling access to sensor data, kinematic states, and chassis actuation.

The hybrid digital twin layer runs on a workstation hosting a CARLA-SUMO co-simulation environment. A workstation-deployed middleware bridges the physical and simulated domains by injecting real vehicle states to synchronize a shadow vehicle in simulation and by encoding decision/motion-planning commands into executable chassis commands for closed-loop control.

The V2X link is realized by an OBU-RSU transparent forwarding chain: the OBU is equipped on the vehicle, and the RSU is connected to the workstation, providing a low-latency bidirectional channel for real-time physical-virtual interaction.

\subsection{ End-to-End Closed-Loop}

\begin{figure}[t]  
    \centering
    \includegraphics[width=\linewidth]{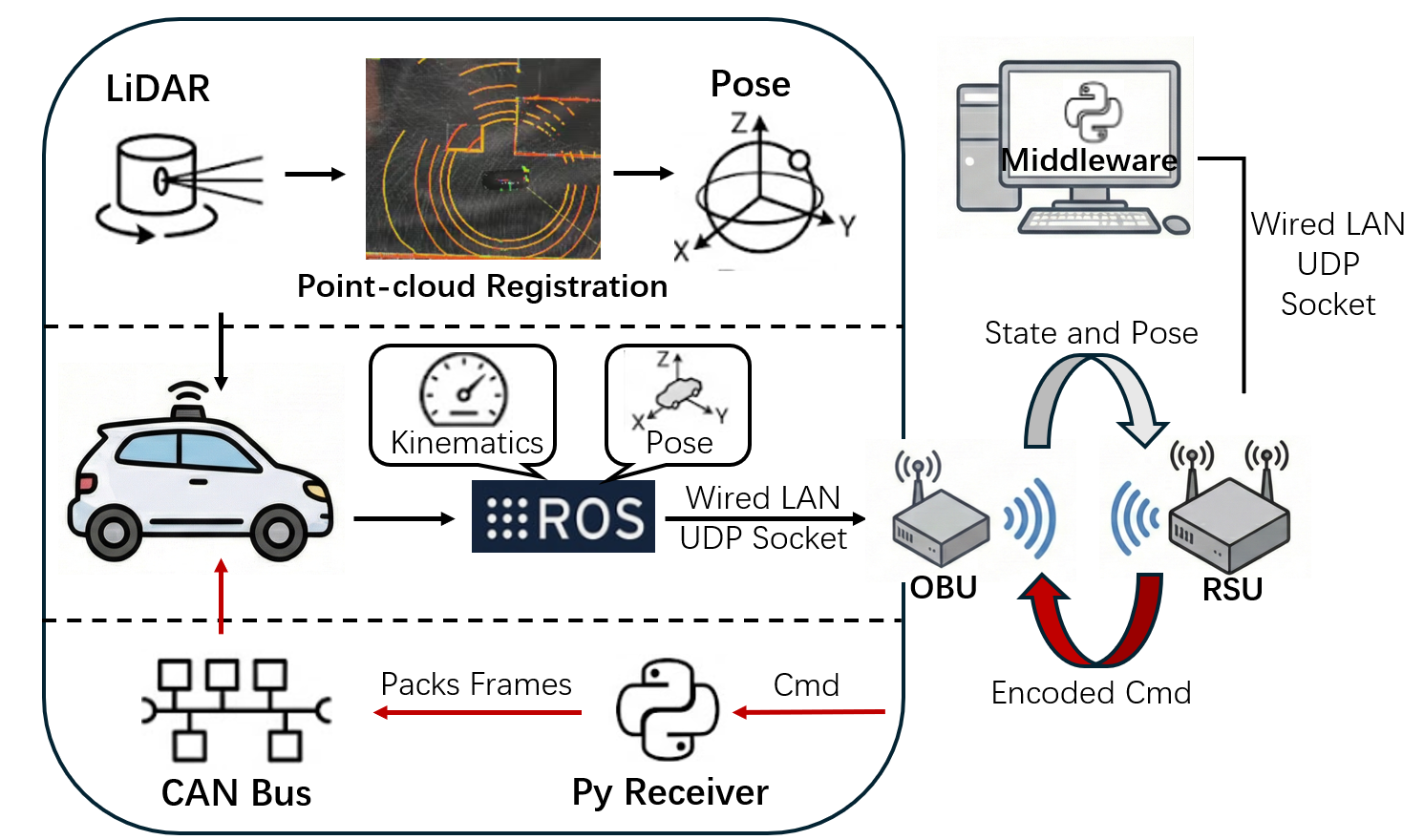} 
    \caption{Architecture of the end-to-end closed-loop.}
    \label{fig:closeloop}
\end{figure}

\begin{figure}[t]  
    \centering
    \includegraphics[width=\linewidth]{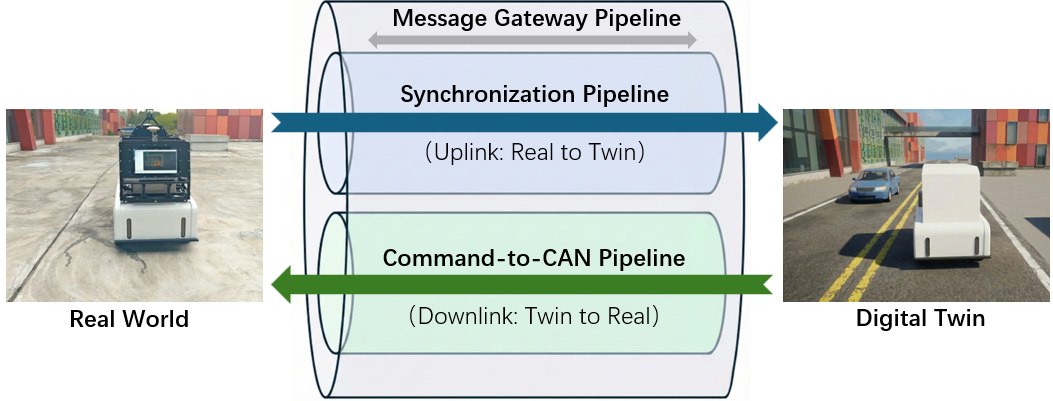} 
    \caption{Three pipelines of the middleware.}
    \label{fig:3pipelines_single}

\end{figure}

The proposed framework establishes an end-to-end closed-loop that couples the physical vehicle with the hybrid twin world through a low-latency V2X communication link, as shown in Fig.~\ref{fig:closeloop}. On the vehicle side, the on-board computer subscribes to Robot Operating System (ROS) topics to obtain kinematic states and the vehicle pose in the test-site coordinate frame. Prior to experiments, the vehicle is driven to scan the site and build a LiDAR point cloud map; online localization is then performed via point cloud registration to provide consistent global coordinates during testing. 

At runtime, the ROS-derived state and pose are serialized into compact UDP packets and transmitted over a wired Local Area Network (LAN) to the OBU. The OBU forwards these upstream packets transparently to the RSU, which is connected via a similar wired link to the workstation hosting the CARLA-SUMO co-simulation. The workstation continuously receives upstream data and uses it to drive real-time updates of the corresponding vehicle entity in the twin world.

For the control return path, commands generated on the workstation are sent back to the vehicle through the same OBU-RSU forwarding channel. A lightweight Python receiver running on the vehicle packs the incoming command frames and transmits them onto the chassis CAN bus for actuation. This bidirectional pipeline enables real-time state upload and command feedback, forming a closed-loop testing path between the physical vehicle and the hybrid twin environment.

\subsection{Hybrid Digital Twin Middleware }

A fundamental design in our mixed-reality setup is to execute control on the physical vehicle and synchronize only the measured kinematic state into the simulator. Compared with the alternative control-in-simulation-then-mirror-to-real strategy, this design has two practical benefits. (i)~The end-to-end delay is dominated only by state transport (V2X) and can be kept below 50 ms. (ii)~It preserves physical realism: the real vehicle naturally obeys inertia and actuator limits, while a simulated vehicle can exhibit unrealistically instantaneous acceleration or steering, which would require additional prediction or compensation methods to reconcile the virtual-physical gap \cite{wang2017parallel}. By treating the physical vehicle as the time reference and letting the simulator follow, the twin world remains anchored to real dynamics.

To leverage this high-fidelity synchronization as a testbed for CAV verification, the middleware functions as the framework's core component, managing the bidirectional data flow required for closed-loop evaluation. On the uplink, it synchronizes the shadow vehicle by mapping the real vehicle's kinematic state into the simulation. On the downlink, it captures control commands generated in the twin world and transmits them to the real vehicle's chassis for immediate execution. To support this workflow, the middleware implements three core functionalities:
\begin{itemize}
\item Message Gateway: It serves as an information relay between the physical vehicle and the hybrid twin environment.
\item Shadow Vehicle Synchronization: It maintains a digital twin by mapping high-rate kinematic updates from the real vehicle into the simulation.
\item Command-to-CAN Conversion: It encodes high-level motion planning commands into CAN frames compliant with the chassis communication matrix for direct execution.
\end{itemize}
Accordingly, we structure the middleware into three distinct pipelines, as illustrated in Fig.~\ref{fig:3pipelines_single}.
\subsubsection{Message Gateway Pipeline}
The Message Gateway Pipeline is implemented as a distributed module deployed on (i)~the vehicle on-board computer and (ii)~a workstation, enabling bidirectional information exchange between the physical world and the twin world. Since the communication devices are directly connected by wired links (the OBU is connected to the vehicle via Ethernet, and the RSU is connected to the workstation via Ethernet), the two sides form dedicated local area networks. Under this setting, we adopt UDP sockets for message transport: the typical drawbacks of UDP (e.g., congestion-induced loss, retransmission absence, and fairness issues) are substantially mitigated. In our measurements, the link exhibits high throughput, low latency, and nearly no packet loss or congestion, which makes UDP a practical choice for real-time closed-loop synchronization.

V2X communication is inherently broadcast-oriented and shares the same wireless spectrum across multiple transmitters, which introduces unrelated traffic noise at the receiver. However, our system only requires the transparent relayed payloads between the paired OBU/RSU. To reliably isolate the desired stream, the sender prepends a unique marker header to every UDP datagram, and the receiver performs lightweight filtering by validating this header before parsing. This design provides an efficient message identification and filtering scheme without introducing additional handshake overhead.

On the vehicle side, the gateway periodically extracts the latest localization and motion states (e.g., position, yaw, velocity, and yaw rate), serializes them into a compact structured representation, and transmits them to the OBU for transparent forwarding. On the workstation side, the gateway receives the relayed datagrams, verifies the marker, decodes the structured fields, and updates a thread-safe state store to support hybrid twin synchronization and downstream actuation. Our tests show, the end-to-end communication latency of this pipeline is below 50 ms, which satisfies the timing requirements of our real-time digital twin loop.

\begin{figure}[t]  
    \centering
    \includegraphics[width=\linewidth]{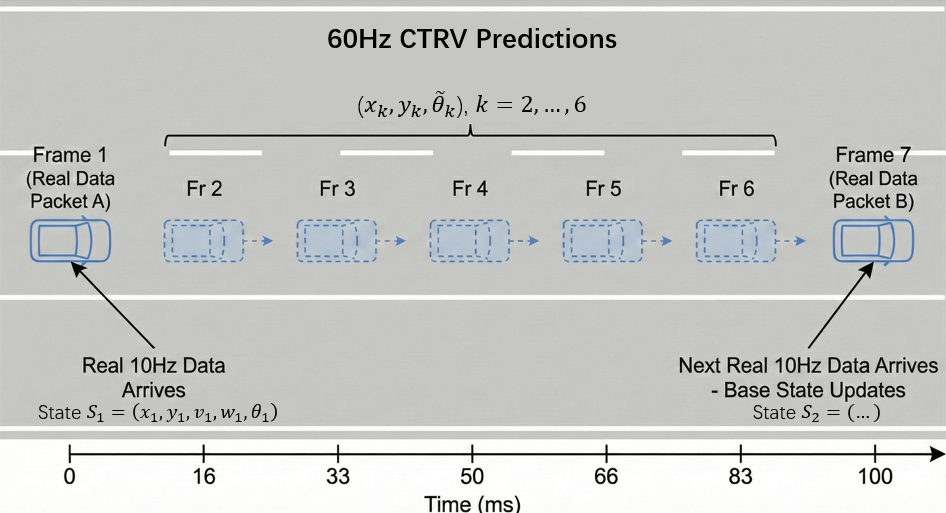} 
    \caption{Visualization of state propagation between discrete updates via CTRV.}
    \label{fig:ctrv}
    
\end{figure}

\subsubsection{Shadow Vehicle Synchronization Pipeline}
This module is a core component of the middleware and is critical for realizing a high-fidelity twin world. Hybrid digital twin testing is challenging for three primary reasons. First, the coordinate conventions between the real site and the simulator are not inherently identical (origin placement, axis directions, yaw definition). Second, the physical world and the simulation world do not share an identical time flow: simulator stepping is influenced by rendering load and runtime scheduling, whereas the physical vehicle evolves continuously in wall-clock time. Third, high-fidelity synchronization must explicitly account for both the transport latency of state transmission and the discrete vehicle state updates limited by the physical or mechanical characteristics of on-board sensors. Our synchronization pipeline is designed around these constraints and aims to produce temporally aligned and visually smooth motion of a shadow vehicle in simulators that mirrors the real vehicle as closely as possible.

Since the simulation environment is reconstructed at full scale with the real test site, no additional scaling is needed for spatial alignment. As a result, real-to-virtual mapping only requires a lightweight coordinate-frame conversion: we apply a planar rotation and translation to align the origin and axis directions between the two worlds, and we use a consistent heading convention with an optional sign flip if the coordinate handedness differs. This simple transform is sufficient to project the physical vehicle's measured position and heading into the simulation world with one-to-one spatial correspondence.

With spatial alignment established, the core of achieving high-fidelity synchronization lies in employing a latency-aware forward prediction mechanism that simultaneously compensates for transmission delays and mitigates the motion discontinuities caused by the discrete updates. Specifically, our CAV transmits kinematic states at a fixed frequency (normally 10\,Hz). Directly mapping these discrete packets to a high-frame-rate simulation (e.g. 60\,Hz in CARLA) would result in discontinuous motion artifacts. To mitigate this, we employ a continuous extrapolation mechanism based on the Constant Turn Rate and Velocity (CTRV) model. For every rendering frame $k$ at simulation time $t_{k}$, the vehicle pose is propagated based on the time horizon $\Delta t$. This horizon is calculated relative to the timestamp $t_{pkt}$ of the latest state packet, augmented by a calibrated lead margin $\delta$ to compensate for residual latencies:

\begin{equation}
\label{eq:time_delta}
\Delta t = t_{k} - t_{pkt} + \delta
\end{equation}

\noindent Given this $\Delta t$, the state is updated via the CTRV model:

\begin{equation}
\label{eq:ctrv}
\begin{aligned}
x_{k} &= x_{t} + \frac{v_{t}}{\omega_{t}} \left( \sin(\theta_{t} + \omega_{t} \Delta t) - \sin(\theta_{t}) \right) \\
y_{k} &= y_{t} - \frac{v_{t}}{\omega_{t}} \left( \cos(\theta_{t} + \omega_{t} \Delta t) - \cos(\theta_{t}) \right) \\
\tilde{\theta}_{k} &= \theta_{t} + \omega_{t} \Delta t
\end{aligned}
\end{equation}

\begin{figure*}[t] 
    \centering
    \includegraphics[width=1.0\textwidth]{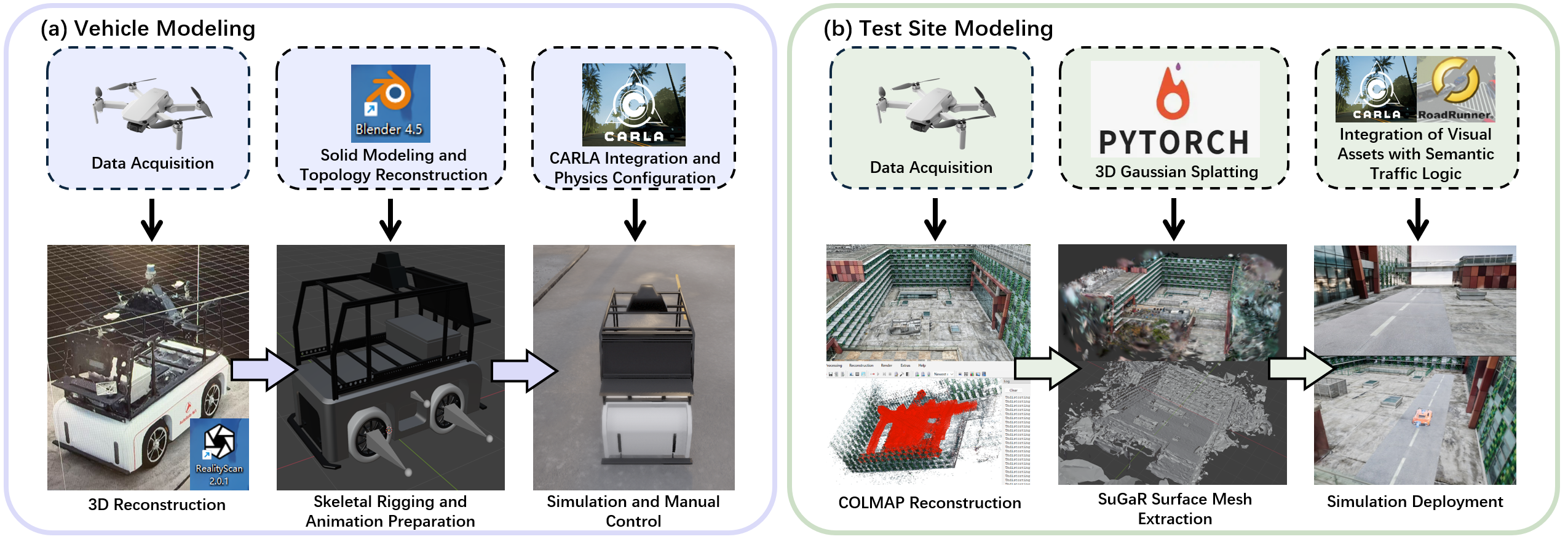}
    \caption{High-Fidelity modeling pipelines. (a) Vehicle modeling pipeline. (b) Test site modeling pipeline. }
    \label{fig:modeling}
    \vspace{-10pt}
\end{figure*}

\noindent \noindent where $(x_t, y_t, \theta_t)$ and $(v_t, \omega_t)$ denote the 2D pose (position and heading) and kinematic rates (velocity and yaw rate) extracted from the physical vehicle's state packet $S_t$, respectively. The tuple $(x_k, y_k, \tilde{\theta}_k)$ represents the extrapolated pose for the current simulation frame $k$. This extrapolation effectively fills the temporal gaps between the discrete 10\,Hz updates, as shown in Fig.~\ref{fig:ctrv}.

However, raw sensor estimation often introduces high-frequency noise. Standard linear filtering is inapplicable to heading angles due to the periodicity of the angular domain (i.e., the discontinuity at $\pm \pi$). To address this, we define a wrapping operator $\mathcal{W}(\phi) = ((\phi + \pi) \bmod 2\pi) - \pi$ that maps angles to the range $(-\pi, \pi]$. We then apply a Riemannian-manifold-aware Exponential Moving Average (EMA) to suppress jitter while preserving topological consistency:

\begin{equation}
\label{eq:ema}
\hat{\theta}_{k} = \mathcal{W}\left( \hat{\theta}_{k-1} + \alpha \cdot \mathcal{W}(\tilde{\theta}_{k} - \hat{\theta}_{k-1}) \right)
\end{equation}

\noindent where $\hat{\theta}_{k}$ is the final smoothed heading, $\tilde{\theta}_{k}$ is the raw extrapolated heading from \eqref{eq:ctrv}, and $\alpha \in [0, 1]$ is the smoothing factor. The inner $\mathcal{W}(\cdot)$ computes the shortest signed angular distance on the unit circle, ensuring that the filter correctly handles transitions across the singularity (e.g., $179^\circ \rightarrow -179^\circ$).

\subsubsection{Command-to-CAN Conversion Pipeline}

\begin{table}[t]
\caption{CAN command message encoding}
\label{tab:can_encoding_prototype}
\centering
\scriptsize
\setlength{\tabcolsep}{3.5pt} 
\renewcommand{\arraystretch}{1.15}
\begin{tabular}{l l c c c c c}
\toprule
Msg & Signal & Start Byte & Start Bit & Len (bit) & Type & Offset \\
\midrule

\multirow{1}{*}{IECU\_Flag} 
& IECU\_Flag 
& 0 & 0  & 8  & uint8  & 0 \\
\midrule

\multirow{2}{*}{IECU\_Steer} 
& Steer\_Valid 
& 0 & 0  & 8  & uint8  & 0 \\
& Steer\_AngleCmd 
& 4 & 32 & 16 & uint16 & -30.0 \\
\midrule

\multirow{4}{*}{IECU\_Speed} 
& Speed\_Valid 
& 0 & 0  & 8  & uint8  & 0 \\
& WorkMode 
& 2 & 16 & 8  & uint8  & 0 \\
& Gear 
& 3 & 24 & 8  & uint8  & 0 \\
& AccelCmd 
& 4 & 32 & 8  & uint8  & -5.0 \\
\midrule

\multirow{2}{*}{IECU\_Brake} 
& Brake\_Valid 
& 0 & 0  & 8  & uint8  & 0 \\
& BrakeCmd 
& 1 & 8  & 8  & uint8  & 0 \\
\midrule

\multirow{3}{*}{Light\_Flag} 
& TurnLeft 
& 0 & 0  & 1  & uint1  & 0 \\
& TurnRight 
& 0 & 1  & 1  & uint1  & 0 \\
& BrakeLight 
& 1 & 8  & 8  & uint8  & 0 \\
\bottomrule
\end{tabular}
\end{table}

This pipeline realizes \emph{execute control on the physical vehicle}, enabling decision/motion-planning policies to be developed and executed in the twin world while being applied to the real platform through the chassis CAN interface. Compared with an alternative we do not adopt---injecting twin-world information into the physical vehicle sensors and deploying the full decision/motion-planning stack on-board \cite{li2024seamless}---our design is lower-latency (direct actuation interface without sensor-level synthesis/injection), and easier to deploy (no modification of on-board perception stack). These properties make the pipeline particularly suitable for CAV verification, where rapid iteration and repeatable execution of diverse controllers are needed.

We implement a lightweight \emph{command-to-CAN} translator according to the vehicle chassis CAN communication matrix. The pipeline accepts control commands from either remote driving or an off-board decision/motion-planning algorithm (e.g., acceleration, braking, steering, and lighting), quantizes them into integer fields, and packs them into fixed-length CAN frames. As summarized in Table~\ref{tab:can_encoding_prototype}, all messages use an 8-byte payload with Intel (little-endian) packing. Specifically, the \texttt{IECU\_Flag} message serves to engage the chassis control interface. For the primary control messages (\texttt{IECU\_Steer}, \texttt{IECU\_Speed}, and \texttt{IECU\_Brake}), the zeroth byte acts as a validity indicator. The actuation commands are mapped to specific offsets: the steering angle is packed into Byte~4 of \texttt{IECU\_Steer}, the acceleration demand into Byte~4 of \texttt{IECU\_Speed}, and the braking intensity into Byte~1 of \texttt{IECU\_Brake}. Notably, we configure the \texttt{WorkMode} field in \texttt{IECU\_Speed} to explicitly accept acceleration values (rather than throttle pedal percentages) and fix the transmission gear to the forward drive position.

Overall, any control-related instruction, regardless of whether it originates from a remote operator or an off-board autonomy stack, can be compiled by the Command-to-CAN Conversion Pipeline into chassis-executable CAN frames, thereby bridging twin-world policy generation and physical-vehicle execution in a deployment-friendly and low-latency manner.

\subsection{ High-Fidelity Modeling of the Vehicle and Test Site  }
High-fidelity, scale-consistent modeling reduces the sim-to-real gap in digital twin vehicle-in-the-loop experiments. Metrically accurate geometry preserves correct distances, occlusions, and collision boundaries, while consistent scale improves the realism of sensor simulation and ensures fair evaluation. In the following, we describe the modeling pipelines for the vehicle (dynamic vehicle model) and the test site (static traffic environment), respectively, as illustrated in Fig.~\ref{fig:modeling}.

\noindent\textbf{Vehicle modeling}: To obtain a simulation asset that closely matches the real vehicle, we adopt a photogrammetry-based multi-view modeling pipeline and convert the reconstructed surface model into a physics-ready, controllable CARLA vehicle. Specifically, a Unmanned Aerial Vehicle (UAV) performs a surround-flight around the vehicle and records 4K video from diverse heights and viewing angles to ensure sufficient texture and geometric coverage. The video is processed in RealityScan 2: representative frames are extracted and used for photogrammetric reconstruction to generate a dense point cloud and an initial textured mesh. Since the reconstructed mesh is a surface shell and is not directly suitable for dynamics simulation, we export it to Blender for solid reconstruction and topology refinement, where a watertight body is rebuilt with proper polygonal structure by referencing the scanned mesh. The final model is then rigged by separating the four wheels and binding the body and wheels to an armature, and exported as FBX. In CARLA, we configure collision primitives (e.g., body box and wheel spheres), calibrate wheel/tire dimensions according to the modeled scale, and implement basic wheel-rotation animation via an animation blueprint. Finally, the blueprint is registered and assigned to the shadow vehicle in the twin world, so that the simulated counterpart matches the physical vehicle in a one-to-one manner in both geometry and scale for high-fidelity experiments.

\noindent\textbf{Test site modeling}: To construct a photorealistic and physically interactive digital twin of the test site, we implement a pipeline integrating neural rendering with explicit geometric extraction. Specifically, the process begins with systematic aerial photogrammetry, where a UAV captures high-resolution 4K imagery with 70\% overlap to ensure robust feature correspondence for Structure-from-Motion via COLMAP. To achieve high visual fidelity, the scene is reconstructed as a neural radiance field using 3D Gaussian Splatting (3DGS), which accurately captures complex lighting and surface properties \cite{kerbl20233d}. Since 3DGS lacks the explicit topology required for collision detection, we employ the SuGaR algorithm \cite{guedon2024sugar} to extract a surface-aligned triangular mesh from the radiance field. This raw geometry undergoes industrial-grade refinement using Reconstruction Master to repair non-manifold artifacts and smooth terrain gradients, ensuring realistic suspension interaction. Finally, we utilize RoadRunner to synthesize the semantic traffic logic (OpenDRIVE) and import both the optimized static mesh and navigation rules into CARLA, establishing a standardized simulation map that unifies visual realism with physical validity.

\subsection{Performance Verification}
To quantify real-to-virtual synchronization, we measure the instantaneous position error between the real vehicle and the synchronized shadow vehicle at representative time instants:
\begin{equation}
e_p(t)=\sqrt{(x_r(t)-x_s(t))^2+(y_r(t)-y_s(t))^2},
\end{equation}
where $(x_r(t),y_r(t))$ denotes the real-vehicle position transformed into the CARLA coordinate frame using the same coordinate conversion and offset compensation as in the synchronization pipeline, and $(x_s(t),y_s(t))$ is obtained from the synchronized vehicle actor in CARLA. The measured errors after timestamp matching are summarized in Table~\ref{tab:sync_error}.

\section{System Extension and Validation}

\begin{figure}[t]  
    \centering
    \includegraphics[width=\linewidth]{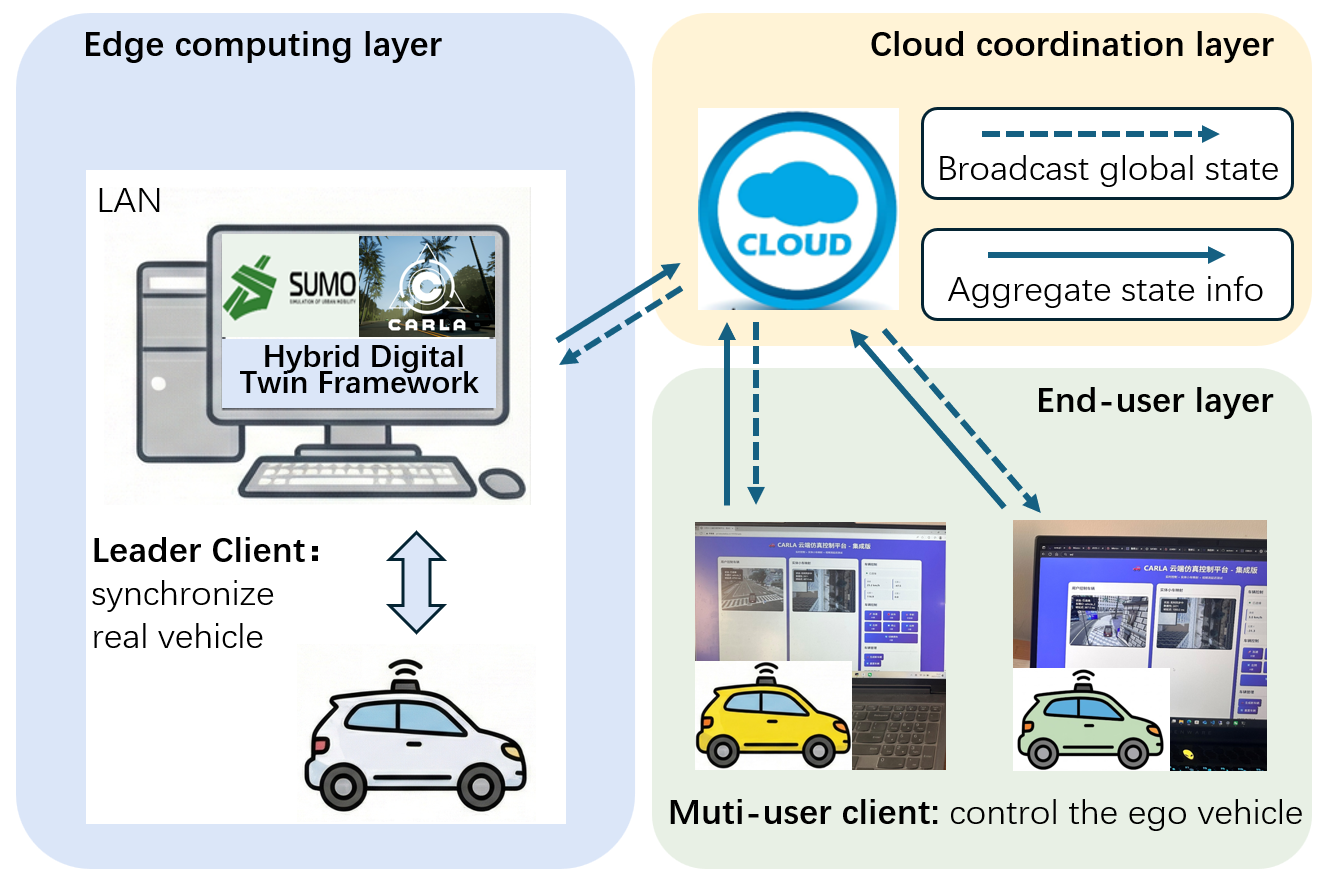} 
    \caption{Cloud-edge collaborative deployment architecture.}
    \label{fig:cloudedge}
\end{figure}

\begin{table}[t]
\centering
\caption{Instantaneous synchronization error at representative time instants}
\label{tab:sync_error}
\renewcommand{\arraystretch}{1.15}
\setlength{\tabcolsep}{3pt}
\begin{tabular}{|c|c|c|c|c|c|c|c|c|}
\hline
$t$ (s) & 1 & 2 & 3 & 4 & 5 & 6 & 7 & 8 \\
\hline
$e_p(t)$ (m) & 0.008 & 0.025 & 0.032 & 0.036 & 0.028 & 0.035 & 0.027 & 0.030 \\
\hline
\end{tabular}
\end{table}

\subsection{Extension for multimodal interactions}
To extend the proposed platform toward multi-vehicle coordination and human-in-the-loop interaction, we implement a cloud-edge collaborative deployment architecture. In this setup, the shadow vehicle synchronized with the physical vehicle can interact not only with background traffic but also with multiple remotely controlled virtual vehicles in a shared simulation world. This enables the construction of diverse and challenging scenarios, such as cut-ins and sudden lane changes, for interactive CAV testing. Meanwhile, the cloud-edge design preserves low-latency local rendering and control while maintaining lightweight global state synchronization for multi-user access.

\subsection{Deployment Architecture and Workflow}

To reduce the latency of cloud-centric digital twin testing, we design a cloud-edge collaborative deployment architecture, as shown in Fig.~\ref{fig:cloudedge}. The system consists of three layers.

\noindent\textbf{Cloud coordination layer}:
A cloud server acts as the relay and coordination center of the system and maintains global consistency via lightweight state exchange. Specifically, it aggregates key state information, including vehicle poses from all clients and background-traffic updates generated by the leader, and continuously broadcasts a unified global state to each participant. 

\noindent\textbf{Edge computing layer}:
A designated Edge server deploys the hybrid digital twin testing framework and maintains a shadow vehicle that is synchronized with the physical vehicle (also called leader client) in real time. The physical vehicle states are collected by the leader-side proxy and encapsulated before being uploaded to the cloud through a stable TCP connection. By broadcasting these states to all participants, the system guarantees that every user observes a consistent representation of the physical vehicle in the shared virtual environment.

\noindent\textbf{End-user layer}:
This layer is designed to provide low-latency interaction for each participant. Each user runs a full CARLA instance locally, where rendering, physics simulation, and control are executed in a local closed-loop to minimize interaction delay. A lightweight local web server serves as a proxy for the user interface and communicates with the cloud via WebSocket. 

\begin{table}[t]
\centering
\caption{Latency Comparison between Two Architectures}
\label{tab:mess-transfer-latency}

\setlength{\tabcolsep}{6pt}
\renewcommand{\arraystretch}{1.1}
\small
\begin{tabular}{|l|c|c|}
\hline
Transfer &
\begin{tabular}[c]{@{}c@{}}Fully\\ Cloud-Centric\end{tabular} &
\begin{tabular}[c]{@{}c@{}}Cloud--Edge\\ Collaborative\end{tabular} \\
\hline
Real Car $\rightarrow$ Local side & \multicolumn{2}{c|}{10--45~ms} \\
\hline
Local $\rightarrow$ Cloud & 50--100~ms & 50--100~ms \\
\hline
Cloud $\rightarrow$ Local & 150--200~ms & 100--150~ms \\
\hline
Total Data Transfer & 400--1000~ms & 100--300~ms \\
\hline
Vehicle Images & Bad & Good \\
\hline
\end{tabular}
\end{table}

\textbf{Workflow}:
Computation-intensive tasks, including high-fidelity shadow-vehicle generation and synchronization, are handled at the edge computing layer, forming an ultra-low-latency local loop near the physical vehicle. Meanwhile, each user periodically uploads lightweight ego states to the cloud, which aggregates and broadcasts the unified global world state so that all local CARLA instances remain synchronized in the shared virtual environment.

\subsection{Latency Evaluation }
To enable a clear comparison with the proposed architecture, we also implement a fully cloud-centric deployment for latency evaluation. In this baseline, all simulators are hosted on the cloud server, and all simulation outputs are streamed from the cloud to the clients throughout the workflow. The measured end-to-end latency results are summarized in Table~\ref{tab:mess-transfer-latency}. Although current evaluation is under ideal conditions, real-world V2X uncertainties (e.g., packet loss, jitter) are important and will be investigated in future work.

\section{Conclusion}
This paper presented a real-time hybrid digital twin testing framework for CAV verification that combines high-fidelity co-simulation with real-vehicle-in-the-loop operation. The framework enables consistent physical-digital synchronization, closed-loop vehicle control, and collaborative testing via a cloud-edge deployment mode, which supports scalable participation while preserving low-latency interaction. Future work will focus on further reducing end-to-end latency, supporting the integration and synchronized representation of multiple physical vehicles in the twin world, enhancing V2X-aware evaluation by incorporating communication simulation, and conducting closed-loop validation of representative CAV functions (e.g., automatic emergency braking and lane change) on the proposed platform.

\bibliographystyle{IEEEtran}
\bibliography{ref}

\end{document}